\documentclass{article}

\usepackage[final]{nips_2017}

\usepackage[utf8]{inputenc} 
\usepackage[T1]{fontenc}    
\usepackage{hyperref}       
\usepackage{url}            
\usepackage{booktabs}       
\usepackage{amsfonts}       
\usepackage{nicefrac}       
\usepackage{microtype}      
\usepackage{amsmath}
\usepackage{epsfig}
\usepackage{graphicx}
\usepackage{floatrow}
\newfloatcommand{capbtabbox}{table}[][\FBwidth]

\usepackage{blindtext}

\title{Cost-Effective Active Learning for Melanoma Segmentation}

\author{
  Marc Gorriz \\ \textbf{Xavier Giro-i-Nieto} \\
  Universitat Politecnica de Catalunya\\
  Barcelona, Catalonia/Spain\\
  \texttt{xavier.giro@upc.edu} \\
  \And
  Axel Carlier\\ \textbf{Emmanuel Faure} \\
  University of Toulouse\\
  Toulouse, France\\
  \texttt{axel.carlier@enseeiht.fr}
}

\begin{document}

\maketitle

\begin{abstract}
We propose a novel Active Learning framework capable to train effectively a convolutional neural network for semantic segmentation of medical imaging, with a limited amount of training labeled data. Our contribution is a practical Cost-Effective Active Learning approach using dropout at test time as Monte Carlo sampling to model the pixel-wise uncertainty and to analyze the image information to improve the training performance. 
\end{abstract}

\section{Motivation}
One of the major problems in medical diagnosis is the subjectivity of the specialist’s decisions. More specifically, in the fields of medical imaging interpretation, the experience of the specialist can greatly determine the outcome of the final diagnosis. Manual inspection of medical images can in some cases be very tedious, time-consuming and subject to errors on part of the interpreter. This has led to a growth of computer vision algorithms to support medical diagnostics. 
The wide success of deep learning in computer vision has increased the accuracy of the predictions, renovating the interest in computer-assisted medical diagnosis.


Nevertheless, deep convolutional neural networks (CNNs) are defined by a huge amount of trainable parameters that require very large amounts of labeled data for training. 
This may be a heavy handicap in the medical imaging field, where the annotation costs of highly qualified professionals  make very challenging the generation of large labeled datasets. Active Learning (AL) is an established approach to reduce this labeling workload in order to select, in an iterative way, a subset of informative examples from a large collections of unlabeled images.
The chosen candidates are labeled and subsequently added to the training set. 

There exist multiple active learning methods that choose which samples are to be annotated by the human experts. 
In this work we explore the uncertainty of the pixel-wise predictions as a selection criterion.

The main contributions of this work are the following:
\begin{itemize}
\item The design and training of a framework for medical imaging semantic segmentation using Convolutional Neuronal Networks (CNN) and the Cost-Effective Active Learning method.
\item The development of information interpreters for medical imaging based on \textit{Monte Carlo Dropout} for the analysis of the intrinsic network distribution.
\end{itemize}

The source code of this work is publicly available at \url{https://marc-gorriz.github.io/CEAL-Medical-Image-Segmentation/}.

\section{Related work}
\subsection{Cost-Effective Active Learning (CEAL) algorithm}
An active learning is an algorithm able to interactively query the human annotator (or some other information source) new labeled instances  from a pool of unlabeled data.  Candidates to be labeled can be chosen with  several methods  based on informativeness and uncertainty of the data. 
Opposite to common Active Learning approaches that only consider the most informative and representative samples, a Cost-Effective methodology \cite{wang2016cost} proposes to automatically select and pseudo-annotate unlabeled samples. A CEAL progressively feeds the samples from the unlabeled dataset into the CNN, and selects two kinds of samples for fine-tuning according to the CNN's output. 
The sampling selection procedure is also known as complementary sample selection. One type of samples are a minority samples with low prediction confidence, which correspond to the most informative/uncertain samples whose prediction score is low. The selected samples are added into the labeled set after an active user (oracle) labels them. Another type of samples are the majority samples with high prediction score, because their prediction is highly confident. For these certain kind of samples, the proposed CEAL automatically assigns pseudo-labels with no human cost. 
These two types of samples are complementary to each other for representing different confidence levels of the current model on the unlabeled dataset.

\subsection{CNN's for Image Segmentation: U-Net architecture}
Our active learning approach is applied over a popular convolutional neural network (CNN) in the medical domain: the U-Net.
The U-Net \cite{ronneberger2015u} is a convolutional neural network architecture designed to solve biomedical image segmentation tasks. It was successfully used for winning the ISBI cell tracking challenge in 2015.
The network combines a convolutional network architecture with a deconvolutional architecture to obtain the semantic segmentation \cite{noh2015learning,krahenbuhl2011efficient}. The convolutional network is composed of a repetitive pattern of two $3$ x $3$ convolutions operations, followed by a ReLU layer and a downsampling process through a $2$ x $2$ maxpooling operation with stride $2$. 
On the other hand, the deconvolutional architecture includes a upsampling operation of the feature map obtained during the contracting path, followed by a $2$ x $2$ deconvolution that fractions the feature map channels into $2$. A posteriori concatenation of the resulting feature map and the obtained during the contracting path is needed, followed by a $3$ x $3$ convolutions and a ReLU layer. The entire network is $23$ convolutional layers deep, where the last layer is used to map each component feature vector related to the number of classes.

\section{Proposed methodology}
\subsection{Image uncertainty estimation}
The active learning criteria used to perform the complementary sample selection is based on the intrinsic distribution of the unlabeled data, ranking the unlabeled pool $D^{U}$ based on their influence on the model. Being $D_{x}^{U}$, $D_{y}^{U}$ the unlabeled data and their labels respectively, we are interested in finding the posterior network distribution $p(W |D_{x}^{U}, D_{y}^{U})$. In general, this posterior distribution is not tractable, therefore we need to approximate the distribution of these weights using variational inference \cite{graves2011practical}. This technique allows us to learn the distribution over the network’s weights, $q(W)$, by minimizing the \textit{Kullback-Leibler (KL)} divergence between this approximating distribution and the full posterior one. 

In \cite{kendall2015bayesian} it is explored the possibility to use \textit{Monte Carlo Dropout} to approximate $q(W)$ term. The dropout works by randomly deactivating network activations following a \textit{Bernoulli} distribution with a basis probability $p_{d}$ per layer. With all the above, the introduction of \textit{Dropout} during the training prevents overfitting \cite{gal2015bayesian}, while at test time it allows introducing pixel-wise sample uncertainty. Being $I_{x}$ a image pixel, we can estimate the uncertainty of its predicted label $\widetilde{I_{y}}$ computing the variance of $T$ different predictions on the same pixel by the effect of \textit{Dropout} through the network weights. The precision of pixel-wise uncertainty maps will depend on the \textit{Dropout} steps $T$ and the \textit{Dropout} probability $p_{d}$. High $p_{d}$ means high variation of network weights, making difficult a consistent result with a finite number of step predictions. As was shown in \cite{kendall2015bayesian} an optimal $p_{d}$ value is $0.5$ and a maximum precision would be obtained when $T \to \infty$.
\begin{figure}[t]
\begin{center}
\includegraphics[width=0.8\linewidth]{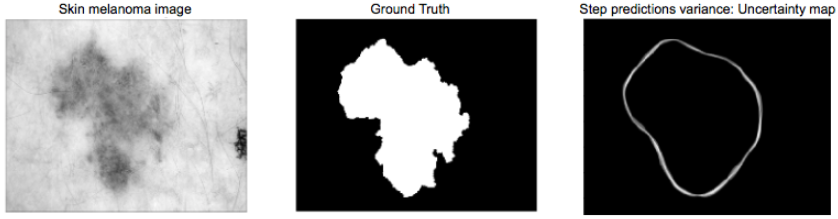}
\end{center}
\caption{Pixel-wise uncertainty map using 10 step predictions.}
\label{fig:fig1}
\end{figure}

In order to integrate the method to the CEAL complementary sample selection, the pixel-wise uncertainty must be transformed into a numerical score to estimate the prediction confidence. We propose summing up all the pixel values from the uncertainty map, obtaining this way higher scores for those most doubtful segmentations. 

However, a simple pixel-wise addition does not cope with the variability of predictions across the contours. 
The further from the contour, the greater should be the contribution to the overall uncertainty of the prediction.
For this reason, we introduce an additional weighting step based on the distance map over the predicted segmentation, which corresponds to computing the euclidean distance of each pixel to the closest pixel of the contour \cite{breu1995linear}.
This distance map is multiplied with the uncertainty map, so that the values of the further pixels from the predicted contour are boosted.
Intuitively, this processing maker thicker the thickest contours in the uncertainty maps, and thinner the thinnest ones.

\subsection{Complementary sample selection}
\begin{figure}[h]
\begin{center}
\includegraphics[width=1\linewidth]{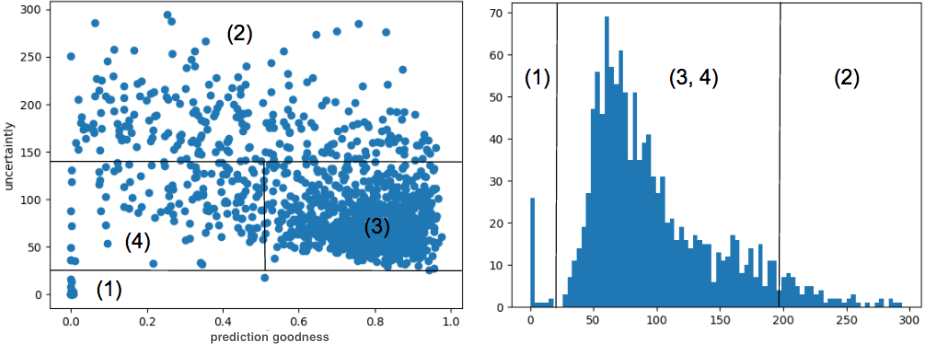}
\end{center}
\caption{Right: Regions uncertainty representation. Left: histogram projection.}
\label{fig:regions}
\end{figure}

Once the uncertainty score is defined, we can visualize its relation with the accuracy of the predictions.
Figure \ref{fig:regions} (left) shows the correlation between both, using as predictor the best model found in the next Section \ref{sec:results}.
The plot allows distinguishing four types of samples:
\begin{enumerate}
\item Undetected melanomas: although they have low uncertainty, they are also the most informative candidates to be manually annotated because they correspond to missing detections but with high certainty.
\item Highly uncertain samples: secondary candidates to be annotated by the oracle.
\item Certain samples: the most common ones, best candidates to be selected as a pseudo-labels.
\item Uncertain and wrong predictions: the most undesirable case whose population aims at be reduce by iterating the active learning algorithm. 
\end{enumerate}

Notice that in the real case scenario, the accuracy of the predictions cannot be computed because the ground truth of the analyzed samples is not available.
In a real case only the uncertainty values can be estimated. 
A histogram of these uncertainty values is depitced in \ref{fig:regions} (right), which corresponds to a projection of \ref{fig:regions} (left) over the uncertainty axis.  
In this case, samples between the third and fourth regions become shuffled and cannot be discriminated. 
With all the above, the complementary data selection method will plan a strategy to select sequentially the correct proportion of samples for each region in each iteration, to improve gradually the network performance, without falling into over-fitting. 

\section{Results and Future work}
\label{sec:results}
All the tests were done with the ISIC 2017 Challenge dataset for Skin Lesion Analysis
towards melanoma detection \cite{gutman2016skin}.
Each image in this dataset has a pixel-wise segmentation of the melanomas. 
In order to simulate an active learning scenario, our experiments only used a small portion of available annotations.
The amount of segments used for training the CNN would increase based on the decisions of the proposed active learning approach, as if they were generated online by human annotators.



The dataset contains $2,000$ RGB dermoscopy images manually annotated by medical experts, by tracing the lesion boundaries in the form of a binary mask (International Skin Imaging Collaboration). The dataset was modified for this work, transforming the original images to gray scale and modifying their aspect ratio to fit the CNN input dimensions. Before starting the interactive learning process, the training sets are initialized based on the Cost-Effective Active Learning methodology. First, it is randomly selected the labeled amount $D^{L}$ from the dataset and then the other labels are deleted initializing the unlabeled set $D^{U}$. 

The CNN was trained by applying data augmentation over the training samples as in \cite{krizhevsky2012imagenet}, to improve the generalization of the model. 

\begin{figure}[h]
\begin{floatrow}
\includegraphics[width=0.60\linewidth]{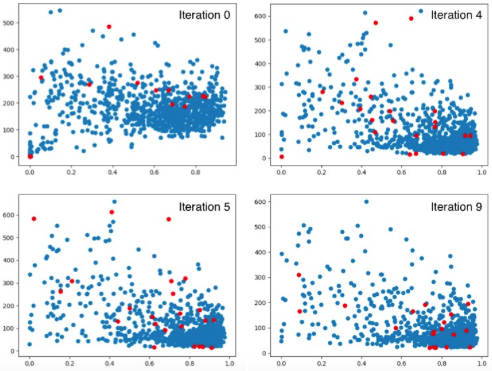}{%
}
\capbtabbox{%
  \begin{tabular}{ll}
    \toprule
    \multicolumn{2}{c}{Initial database balancing}                   \\
    \cmidrule{1-2}
    Initial Set	& Samples	   \\
    \midrule
    Labeled Set	& 600    \\
    Unlabeled Set & 1000   \\
    Test Set    	& 400    \\
    \bottomrule
    \\
    \toprule
    \multicolumn{2}{c}{Sample selection per iteration}                   \\
    \cmidrule{1-2}
    Human annotations	& Samples	   \\
    \midrule
    No-detections	& 10    \\
    Most uncertain  & 10   \\
    Random       	& 15    \\
    \cmidrule{1-2}
    Pseudo annotations	& Progressive	   \\
    \bottomrule
    \\
 
  \end{tabular}
}{%
  
}
\end{floatrow}
\caption{Left: Regions evolution. Red samples showing the manual annotations. Right: Cost-Effective Active Learning data balance and heuristic parameters.}%
\label{fig:fig2}
\end{figure}

The selection of samples in each iteration of the active learning loop was made according to the heuristic parameters summarized in the tables of Figure \ref{fig:fig2}.
The iterative algorithm starts by training a CNN with 600 samples and iteratively selects among a pool of 1,000 images which ones are to be labeled and added to the training set. In each iteration, the proposed algorithm samples 10 images where among those where no melanoma was detected, 10 more images among those with a highest uncertainty score, and 15 more images selected randomly.
Also, those predictions with a confidence score over a certain threshold was selected as pseudo-labels and also included in the training set.
Figure \ref{fig:fig2} also depicts the selected samples at some iterations 0, 4, 5 and 9.


The quality of the segmentations was assessed based on the Dice Coefficient as in \cite{ronneberger2015u}, which is presented in Equation \ref{eq:sbd}:

\begin{equation}
\label{eq:sbd}
\begin{gathered}
\begin{aligned}
\mathrm{DICE}(A,B) =& \frac{2|A\hat{B}|}{|A| + |B|}
\end{aligned}
\end{gathered}
\end{equation}

where $A$ represents the predicted masks for image and  $B$ its ground truth mask.

The results reached a Dice Coefficient \cite{ronneberger2015u} of 74 \% after 9 active learning iterations with 2 training epochs over the CNN per run. Results indicate that after these 9 iterations there were still samples in the central region (4) that may limit the performance of the model. 

\section*{Acknowledgements}
\label{sec:acknowledgements}

This research was supported by contract SGR1421 by the Catalan AGAUR office.
The work has been developed in the framework of project TEC2016-75976-R, funded by the Spanish Ministerio de Economia y Competitividad and the European Regional Development Fund (ERDF). The authors also thank NVIDIA for generous hardware donations.

\small

\bibliographystyle{plainnat}
\bibliography{egbib}

\end{document}